\documentclass{article} % For LaTeX2e
\usepackage{iclr2027_conference,times}

\usepackage{amsmath,amsfonts,bm}

\def\eqref#1{equation~\ref{#1}}
\def\1{\bm{1}}

\DeclareMathAlphabet{\mathsfit}{\encodingdefault}{\sfdefault}{m}{sl}
\SetMathAlphabet{\mathsfit}{bold}{\encodingdefault}{\sfdefault}{bx}{n}

\usepackage{hyperref}
\usepackage{url}
\usepackage{graphicx}
\usepackage[rightcaption]{sidecap}

\graphicspath{{figures/}}

\title{Signal-noise factorization isolates nuisance variation into removable subspaces}

\author{Sakin Kirti \thanks{sakinkirti.github.io} \\
Bioinformatics IDP, Jules Stein Eye Institute \\
University of California, Los Angeles\\
Los Angeles, CA 90095, USA \\
\texttt{sakinkirti@ucla.edu} \\
\And
Joel Zylberberg \thanks{jzlab.org}\\
Jules Stein Eye Institute \\
University of California, Los Angeles \\
Los Angeles, CA, 90095, USA \\
\texttt{joelzy@ucla.edu} \\
}

\iclrfinalcopy % Uncomment for camera-ready version, but NOT for submission.
\begin{document}

\maketitle

\begin{abstract}
Recent theoretical work identified fundamental properties of representational geometry that shape inference ability of deep neural networks. These include signal-noise factorization (SNF), the ability to segregate signal from noise, and signal-signal factorization (SSF), the ability to segregate task-specific and task-irrelevant signals. Here, we built new regularizers that explicitly reinforce these two properties during training. We compared networks trained with these regularizers to $L_2$-regularized baseline networks on the CIFAR-100 image classification task to understand how our new regularizers shape representation geometry and impact performance on a well-known computer vision baseline. Enhancing SNF via regularization improved model performance but enhancing SSF did not. Motivated by biomedical diagnostics applications, we next investigated how our new regularizers affected performance on the BloodMNIST dataset treated with MedMNIST-C corruptions at five severity levels, and found even larger performance gains using the SNF regularizer. To understand the mechanism by which signal-noise factorization produces improved performance, we analyzed the nuisance subspaces across regularization regimes, finding that the SNF-regularized models represent noise in distinct subspaces, separate from class-relevant signal. Because this geometry is explicit, the dominant corruption-induced directions can be estimated on held-out data and projected out of the representations. This manipulation led to a substantial gain in categorization accuracy. These results demonstrate that regularizers that enforce signal-noise factorization can produce substantial improvements on computer vision tasks that contain out-of-distribution image distortions at inference time. They also highlight how explicitly shaping representations affects model performance: isolating nuisance variables from categorical ones is more important than maintaining factorized representations of categorical variables.
\end{abstract}

\section{Introduction}
Deep neural networks learn complex representations across a wide range of domains to solve problems. In most applications, a useful representation must support generalization, transfer, and reliable readout, yet the field is only beginning to understand how the geometric or statistical properties of neural representations affect these outcomes. Rather than asking how to improve performance directly, we ask: how do specific structural properties of hidden representations shape downstream task performance? We answer this by treating regularization as a manipulation to alter two geometric properties identified by a recent theoretical study as the critical ones governing generalization \citep{wakhloo_neural_2026}: signal-signal factorization (SSF), the segregation of task-specific from task-irrelevant signals, and signal-noise factorization (SNF), the segregation of signal from noise. 

Across primate visual cortex, hippocampus, and prefrontal cortex, populations encode task-relevant information in one subspace while segregating task-irrelevant variation into partially orthogonal subspaces \citep{lindsey_factorized_2024, bernardi_geometry_2020, whittington_tolman-eichenbaum_2019}. This factorization appears so consistently that it may constitute a normative principle of  neural coding, and decision-making models with stronger factorization show closer correspondence to neural and behavioral data \citep{lindsey_factorized_2024}, suggesting that preserving distinct sources of variation in separable population directions supports generalization to conditions outside the training distribution.

To this end, we implement two regularizers targeting the SSF and SNF properties. We apply these penalties to the penultimate layer of ResNet-18 networks \citep{he_deep_2015} trained on CIFAR-100 \citep{krizhevsky_cifar-10_2009} and on the biomedically-motivated BloodMNIST classification task \citep{yang_medmnist_2023}, and compare against unregularized and $L_2$-regularized baselines. Beyond measuring accuracy and robustness to corrupted samples, we also analyze the representational structure each regularizer produces, asking not only whether factorization-based regularization helps, but by what mechanism. This analysis reveals that regularizing to improve signal-noise factorization improves categorization performance on samples with out-of-distribution distortions by segregating the effects of the distortions into a subspace of the representation that can be ignored by the readout during inference. For contrast, regularizing to improve signal-signal factorization had little impact on categorization performance.

\subsection{Related Work}

A central question in representation learning is which geometric and statistical properties of hidden-layer activations determine downstream task performance. A substantial body of work characterizes the structure of learned representations after training, by measuring effective dimensionality \citep{abbas_effective_2021}, and the alignment of class-conditional geometry with linear classifiers \citep{papyan_prevalence_2020}. Probing classifiers \citep{belinkov_probing_2021} similarly query frozen representations for task-relevant structure. These methods are diagnostic in that they describe what a representation contains, but do not manipulate it. Our approach is complementary. By introducing targeted regularizers during training, we produce representations with controlled structural properties, enabling a more causal account of how those properties affect readout accuracy.

A parallel line of analytical work asks why gradient descent and weight decay produce the particular solutions they do \citep{neyshabur_search_2014, gunasekar_implicit_2017, arora_implicit_2019, cao_benign_2022, kou_benign_2023}. Neural collapse theory has established that weight decay is not merely incidental but necessary for cross-entropy-trained networks to reach a maximally class-separated, within-class-collapsed geometric optimum \citep{papyan_prevalence_2020, zhu_geometric_2021}, and related work shows that weight decay drives representational isotropy and cross-class orthogonality \citep{kulkarni_disentangling_2026}. Critically, this effect is not monolithic: weight decay appears to act through several distinct mechanisms operating simultaneously \citep{zhang_three_2018, yunis_approaching_2024}. These results are observational, in that geometry is measured in networks trained with weight decay and correlated with performance. We ask instead whether these geometric properties can be optimized explicitly and separately, and in doing so isolate their individual effects on network behavior.

Another class of methods reshape representations through objectives defined on activation statistics. Self-supervised methods such as Barlow Twins \citep{zbontar_barlow_2021} and VICReg \citep{bardes_vicreg_2022} penalize covariance across feature dimensions to prevent representational collapse, targeting label-independent structure. Domain adaptation methods enforce representational invariance across environments, either adversarially \citep{ganin_unsupervised_2015}, through moment matching \citep{sun_deep_2016}, or through invariant-predictor objectives \citep{arjovsky_invariant_2020}. These methods modify representations in service of a downstream objective, like generalization or transfer, without isolating which structural change drives the observed effect. The regularizers considered here are instead designed to manipulate specific, interpretable geometric properties of the class-conditional distribution, so that the link between representation structure and performance can be studied directly.

\section{Factorization-based regularizers}
\label{methods:regularizer_def}

\begin{figure}
    \centering
    \includegraphics[width=\linewidth]{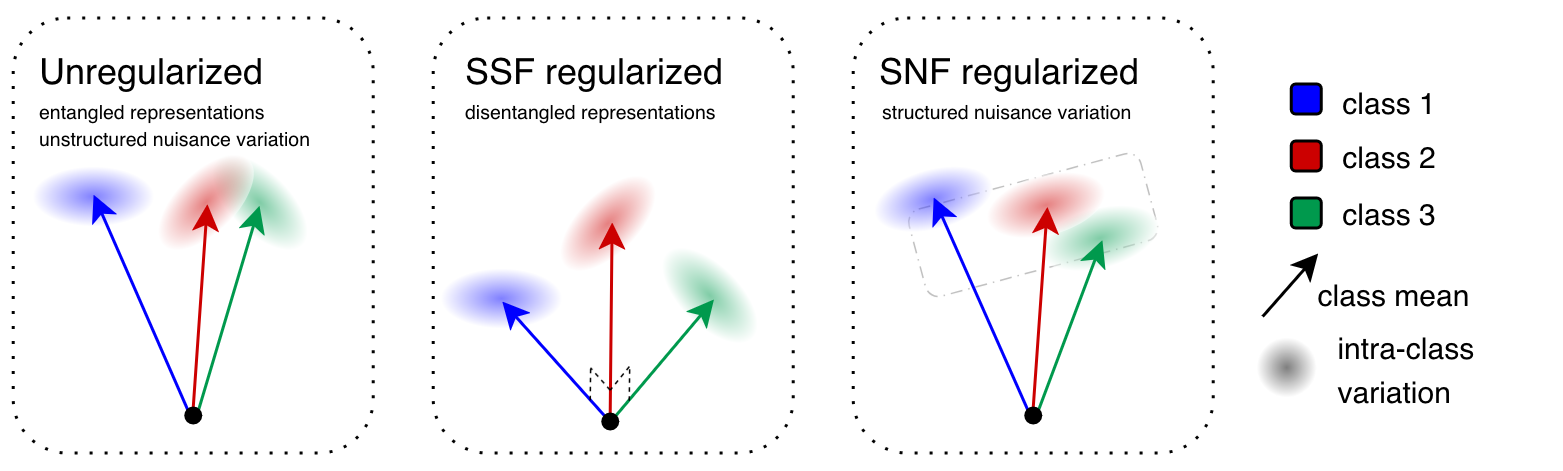}
    \caption{\textbf{Schematic of regularization regimes and their expected effects on representation geometry.}
    (Left) Unregularized networks have entangled representations, with non-orthogonal means and unorganized nuisance variation. (Center) SSF regularization leads to structured representations, with orthogonal class means. (Right) SNF-regularized networks displace nuisance variation into a subspace orthogonal to the class-signal directions, leaving the signal directions unconstrained.}
    \label{fig:overview}
\end{figure}

\citet{wakhloo_neural_2026} derived a lower bound on generalization error as a
function of representational geometry:
\begin{equation}
	E_g = \frac{1}{\pi} \arctan \left( \sqrt{\frac{\pi}{2pc^2 \mathrm{PR}(\Psi)} + \frac{1}{f} + \frac{1}{s} - 1} \right),
\end{equation}
where $p$ is training set size, $c$ is the correlation between population
responses and the task's latent factors, $\mathrm{PR}(\Psi)$ is the effective
dimensionality of the population, and $f$ and $s$ are factorization quantities
defined below. As $p$ grows, $c$ and $\mathrm{PR}(\Psi)$ become asymptotically
unimportant, motivating our focus on $f$ (signal-signal factorization, SSF) and
$s$ (signal-noise factorization, SNF).

\subsection{Preliminaries}
\label{sec:prelim}
Let $Z \in \mathbb{R}^{N \times D}$ denote penultimate-hidden layer activations across $N$ samples and $D$ units, and $Y \in \mathbb{R}^{N \times C}$ the one-hot class labels across $C$ classes. We define the hidden-label, hidden-hidden, and label-label cross-covariance matrices as
\begin{equation}
	\Phi = \frac{1}{N} Z^\top Y \in \mathbb{R}^{D \times C},
    \qquad
    \Psi = \frac{1}{N}Z^\top Z \in \mathbb{R}^{D\times D},
    \qquad
    \Omega = \frac{1}{N}Y^\top Y \in \mathbb{R}^{C \times C}
  \label{eq:covariances}
\end{equation}

Signal-noise factorization, $s$, is defined as
\begin{equation}
  s =\frac{\operatorname{Tr}(\Phi\Phi^\top)^2}
               {\operatorname{Tr}(\Omega)
                \operatorname{Tr}\!\left(
                  \Phi^\top(\Psi - \Phi\Omega^{-1}\Phi^\top)\Phi
                \right)}.
  \label{eq:wakhloo-s}
\end{equation}
The numerator quantifies total signal power, while the denominator scales that power by the noise remaining after the class-predictive component of $\Psi$ is projected out. Higher $s$ indicates that class-signal information occupies a subspace that is well-separated from nuisance variance.

Signal-signal factorization, $f$, is defined as
\begin{equation}
    f=\frac{\operatorname{Tr}(\Phi\Phi^\top)^2}
           {\operatorname{Tr}(\Omega)\operatorname{Tr}(\Phi^\top\Phi\Omega^{-1}\Phi^\top\Phi)}
    \label{eq:wakhloo-f}
\end{equation}
The numerator is identical to the numerator of the $s$ term, denoting total signal power, while the denominator reweights that same signal power according to how class-relevant it is, using the inverse of the label structure as a weighting scheme. Thus, $f$ measures how well the pattern of signal across classes lines up with the underlying class structure itself.

Throughout, we refer to the column space of $\Phi$ as the signal subspace.

\subsection{Signal-signal factorization (SSF)}

The SSF regularization term penalizes deviation of the class-to-class covariance $\Phi^\top\Phi \in \mathbb{R}^{C \times C}$ from a scaled identity matrix:
\begin{equation}
	\mathcal{L}_{f} = \| \Phi^\top\Phi - \lambda I \|_F^2.
    \label{eq:ssf}
\end{equation}
Minimizing $\mathcal{L}_{f}$ encourages class representations to be orthogonal and equal-norm, so that each class is represented independently.

\subsection{Signal-noise factorization (SNF)}

The SNF regularization term is defined as:
\begin{equation}
	\mathcal{L}_{s} = \operatorname{Tr}\left(\Phi^\top (\Psi - \Phi \Omega^{-1} \Phi^\top ) \Phi\right)
    \label{eq:snf}
\end{equation}
where $\Psi - \Phi \Omega^{-1} \Phi^\top$ is the Schur complement \citep{puntanen_schur_2005} of $\Omega$, representing residual activation variance not explained by the signal subspace $\Phi$; minimizing this trace suppresses noise-subspace variance within the signal-coding directions, encouraging activations to be explained by task-relevant structure. $\Omega^{-1}$ is computed explicitly per batch rather than assuming perfectly balanced sampling, which would give $\Omega^{-1} = CI$.

\section{Model training and architecture}
\label{sec:methods}

To evaluate our regularizers across different classification tasks, we use two datasets: CIFAR-100 \citep{krizhevsky_cifar-10_2009}, a standard benchmark enabling comparison to existing literature and confident measurement of model geometry, and BloodMNIST \citep{yang_medmnist_2023}, a benchmark biomedically-motivated classification task with substantial inter- and intra-class nuisance variation.

In both settings, we use a ResNet-18 architecture with CrossEntropy loss, applying our geometric regularizers to the penultimate layer. For each dataset, we perform a joint hyperparameter sweep over $L_2$ strength and SSF/SNF regularization strength, selecting values by validation performance; we compare against unregularized and $L_2$-only baselines, with the $L_2$ strength for these baselines selected the same way. Full training configurations, including optimizer settings, learning rate schedules, augmentation policies, and compute resources, are provided in App.~\ref{app:cifar-training}.

\section{Model and regularizer evaluation}

In this section, we describe the methods used to analyze regularizer impact on model geometry and performance.

\subsection{Model Geometry}
\label{sec:geometry}
To understand the regularizers' effects on model geometry, we compute Wakhloo's $s$ and $f$ (Eq.~\ref{eq:wakhloo-s} and ~\ref{eq:wakhloo-f} respectively) metrics on our model's penultimate activations, evaluating the degree to which the goal of each geometric regularizer is achieved across CIFAR-100-trained models. 

\subsection{Model performance}
We measure whether improvements in representational geometry confer greater performance by first measuring baseline test accuracy on the unseen CIFAR-100 and BloodMNIST test sets. We then measure performance on out-of-distribution examples for CIFAR-100 by measuring each model's ability to transfer performance across corruption strengths by independently training models with weak and strong augmentations and evaluating each model on augmented images of the opposite strength. On the BloodMNIST dataset, we assess the ability to perform on out-of-distribution examples by applying MedMNIST-C corruptions \citep{di_salvo_medmnist-c_2024} to the unseen test set and comparing the performance of each model across corruption severity and corruption type. On the BloodMNIST dataset, we also report the sensitivity and specificity under corruption.

\subsection{Mechanistic subspace analysis} 
To probe the mechanism by which SNF regularization improves performance, we wanted to determine whether nuisance variation was represented in a separate subspace. We extracted 512-dimensional penultimate representations ($Z$) from each trained model on the BloodMNIST test set under clean conditions and all 55 corruption variants (11 types $\times$ 5 severities). We then identified the principal directions of corruption-induced variation as follows. For each model, we formed the perturbation matrix $\Delta=Z_{\mathrm{corrupted}}-Z_{\mathrm{clean}}$ and ran a truncated SVD (top-50 components) on the stacked $\Delta$ across all 55 variants. We then measured the effect of projecting these directions out of the corrupted representations before classification with the frozen linear head, sweeping the projection dimension $k \in \{0, 5, …, 50\}$. To determine whether these subspaces are corruption-specific or general, we additionally computed an $11\times11$ corruption transfer matrix. For each source corruption, a $k$-dimensional subspace was learned from its $\Delta$ and applied to suppress noise in all 11 target corruptions. More detail regarding the analysis is provided in App.~\ref{app:subspace-analysis}.

\section{Results}

\subsection{Representation geometry and CIFAR-100 performance}

\begin{SCfigure}[1.0][t] % The first argument adjusts the width ratio of the caption
    %\centering
    \includegraphics[width=0.66\textwidth]{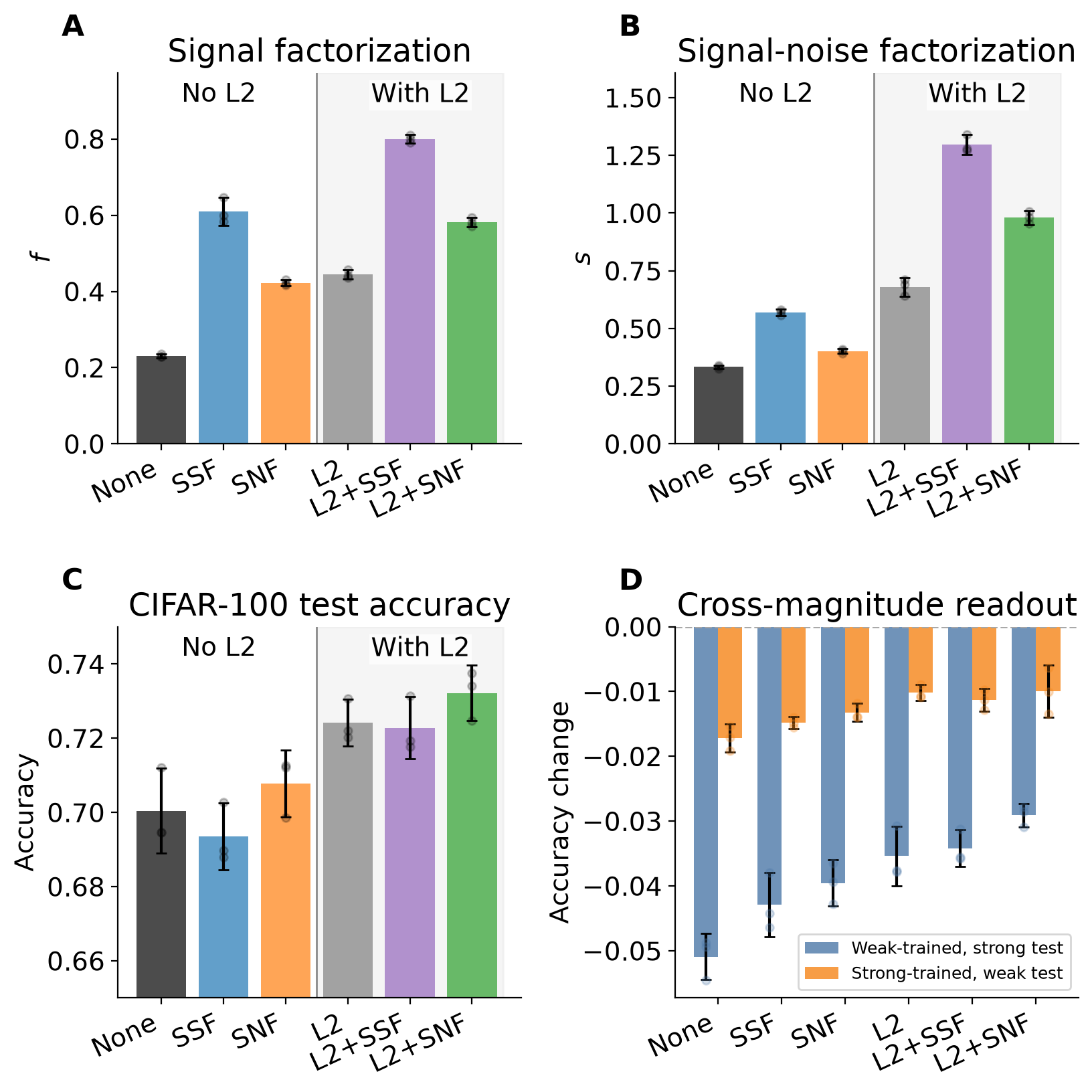}
    \caption{\textbf{Model geometry and CIFAR-100 performance.}
    A) Theoretical $f$ in the penultimate layer, depicting the degree to which class representations are orthogonal. B) Theoretical $s$ in the penultimate layer, depicting the cleanliness of the signal subspace. C) Model performance on CIFAR-100 dataset, demonstrating that SNF regularization leads to small improvements in model performance. D) Difference in classification accuracy when models trained with weak vs strong augmentations are evaluated on weak vs strong augmentations, demonstrating that factorization-based regularizers improve generalization at greater corruption strengths compared to baseline.}
    \label{fig:model_geometry}
\end{SCfigure}

To test our regularizers on a widely used dataset, we train models on the CIFAR-100 dataset \citep{krizhevsky_cifar-10_2009}. We first measure $f$ (Fig.~\ref{fig:model_geometry}a), the theoretical quantity which measures whether class representations occupy orthogonal subspaces. Within both the No-$L_2$ and With-$L_2$ groups, we observe that SSF regularization produces the largest $f$, SNF regularization produces an intermediate increase, and unregularized or $L_2$-only representations show the smallest quantity. This pattern is mirrored in $s$ (Fig.~\ref{fig:model_geometry}b), the theoretical measure of noise in the signal subspace: SSF regularization again maximizes $s$ most aggressively, while SNF regularization provides an intermediate increase relative to no regularization and $L_2$ alone.

We next apply our CIFAR-100-trained models on the held out CIFAR-100 test set (Fig.~\ref{fig:model_geometry}c), where SNF regularization improves performance over no regularization, and when applied jointly with $L_2$, achieves the best overall accuracy among all conditions tested. SSF regularization, in contrast, does not improve generalization and consistently leads to lower accuracy than the unregularized baseline and its $L_2$-paired counterpart. The combination of strong geometric regularization from SSF regularization paired with weaker test performance is consistent with two directions of prior work. First, neural collapse theory demonstrates that in the final layers of training, neural networks already aim to orthogonalize class means, and our results demonstrate that an additional penalty as regularization is unlikely to add to network performance \citep{neyshabur_search_2014, gunasekar_implicit_2017, arora_implicit_2019, cao_benign_2022, kou_benign_2023}. Secondly, while enforcing signal orthogonality via structural loss functions enhances component interpretability, it risks over-constraining the optimization landscape. Real-world signal attributes are rarely perfectly disjoint, and by forcing an orthogonal basis, a model may prioritize structural constraints over data fidelity, producing a rigid solution space that discards subtle, highly discriminative inter-class variation \citep{yan_robust_2025, avants_non-negative_2025}. In contrast, the more moderate geometric effect produced by SNF regularization appears to strike a more useful middle ground.

Finally, to test whether these geometric properties translate into better performance on out-of-distribution data, we trained separate models with weak and strong augmentations ($n=3$ models each) and evaluated each on unseen test data with the opposite augmentation strength (Fig.~\ref{fig:model_geometry}d). Models trained with strong augmentations and evaluated on weak augmentations performed comparably across all regularizers, with accuracy drops of roughly 1-2\% regardless of regularization strategy. In contrast, models trained with weak augmentations and evaluated on strong augmentations showed considerably more variation. The unregularized model degraded most, while SNF regularized and, especially, $L_2$+SNF regularized models degraded the least. This asymmetry suggests that SNF regularization specifically improves the transfer of learned representations to corruption strengths not encountered during training.

\subsection{Regularizers applied to cell type classification}

\begin{figure}[t]
    \centering
    \includegraphics[width=\linewidth]{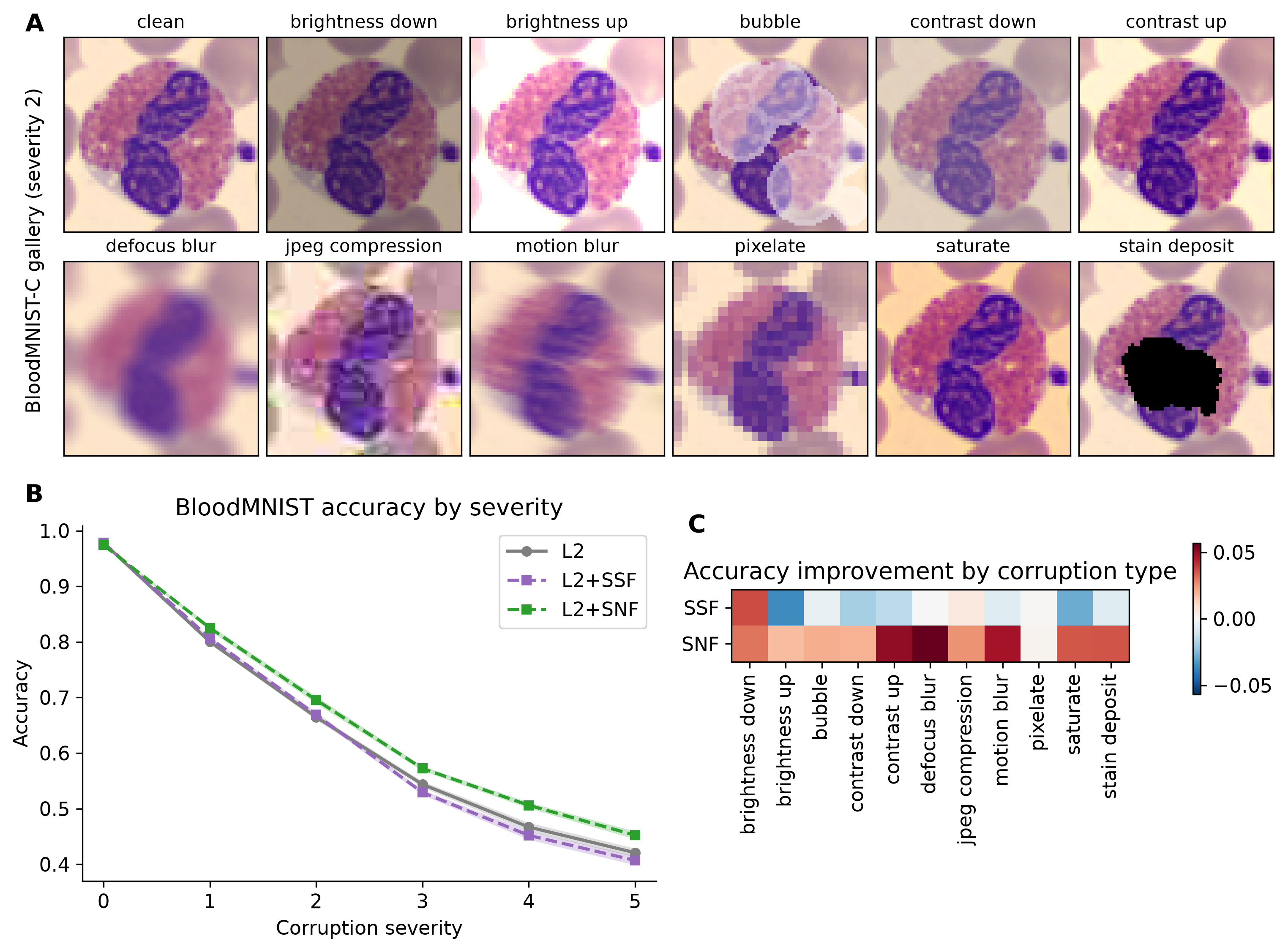}
    \caption{\textbf{Geometric regularizers applied to corrupted medical imaging data.}
    A) Gallery of one example BloodMNIST image with 11 MedMNIST-C corruptions applied at severity 2. B) Comparison of models trained with $L_2$, $L_2$+SSF and $L_2$+SNF regularizers. Plot shows model accuracy and AUC performance on clean test data (severity 0) and on test data with several different corruption severity levels. C) Difference in accuracy per corruption type for models trained with $L_2$+SSF vs $L_2$ and $L_2$+SNF vs $L_2$ regularizers. Red is better performance than $L_2$, blue is worse performance than $L_2$.}
    \label{fig:BloodMNIST_robustness}
\end{figure}

To test whether the benefit of SNF regularization generalizes beyond CIFAR-100, we train a ResNet-18 model on the BloodMNIST benchmarking dataset, an 8-class cytology-based, cell-type classification task with $L_2$ as a shared baseline. We compare this $L_2$-only baseline against $L_2$+SSF (SSF=0.001) and $L_2$+SNF (SNF=0.1), evaluating each model directly on the held-out BloodMNIST test set (3{,}421 images), under both clean and all 11 MedMNIST-C-style corruptions \citep{di_salvo_medmnist-c_2024} at 5 severities each.

On clean test images, all three regularizers are statistically indistinguishable: 97.81$\pm$0.04\% ($L_2$), 97.84$\pm$0.24\% ($L_2$+SSF), and 97.43$\pm$0.13\% ($L_2$+SNF), matching the original MedMNIST baseline of 97.06$\pm$0.11\%. Under corruption, however, $L_2$+SNF regularization separates clearly from both other regularizers at every severity level, while $L_2$+SSF tracks the $L_2$ baseline almost exactly, falling slightly below it at strong severity levels. Averaged over all 11 corruptions and 5 severities, the model with $L_2$+SNF regularization reaches 61.03$\pm$0.83\% accuracy versus 57.93$\pm$0.96\% for $L_2$ regularization alone and 57.27$\pm$0.91\% for $L_2+\mathrm{SSF}$, a $+$3.1\% absolute ($\sim$5.35\% relative) improvement from SNF regularization that comes with no measurable cost to clean test accuracy.

The corruption-specific breakdown (Fig.~\ref{fig:BloodMNIST_robustness}c) shows regularizer performance compared to $L_2$ across corruption types. The picture becomes increasingly clear here, with $L_2$+SNF producing accuracy improvements of up to 5.7\% on defocus blur, with $L_2$+SNF's largest gains concentrated on blur and photometric corruptions. In contrast, $L_2$+SSF shows the opposite pattern, reducing performance across the board. This failure pattern suggests SSF regularization's contrastive class-direction separation does not confer a benefit in corrupted data settings and may over-constrain the network structure \citep{yin_towards_2021}, whereas SNF regularization broadly improves robustness to image distortions. We further breakdown the sensitivity and specificity of models trained with each regularization regime under corrupted data settings in Fig.~\ref{fig:BloodMNIST_sens_spec}. In all classes, regardless of class imbalance (Table~\ref{tab:bloodmnist_data_breakdown}), we see modest improvements to sensitivity, while we see variable changes to specificity per class. SNF regularization, therefore, likely impacts sensitivity to improve overall performance.

\subsection{Mechanistic subspace analysis}

\begin{figure}[t]
    \centering
    \includegraphics[width=\linewidth]{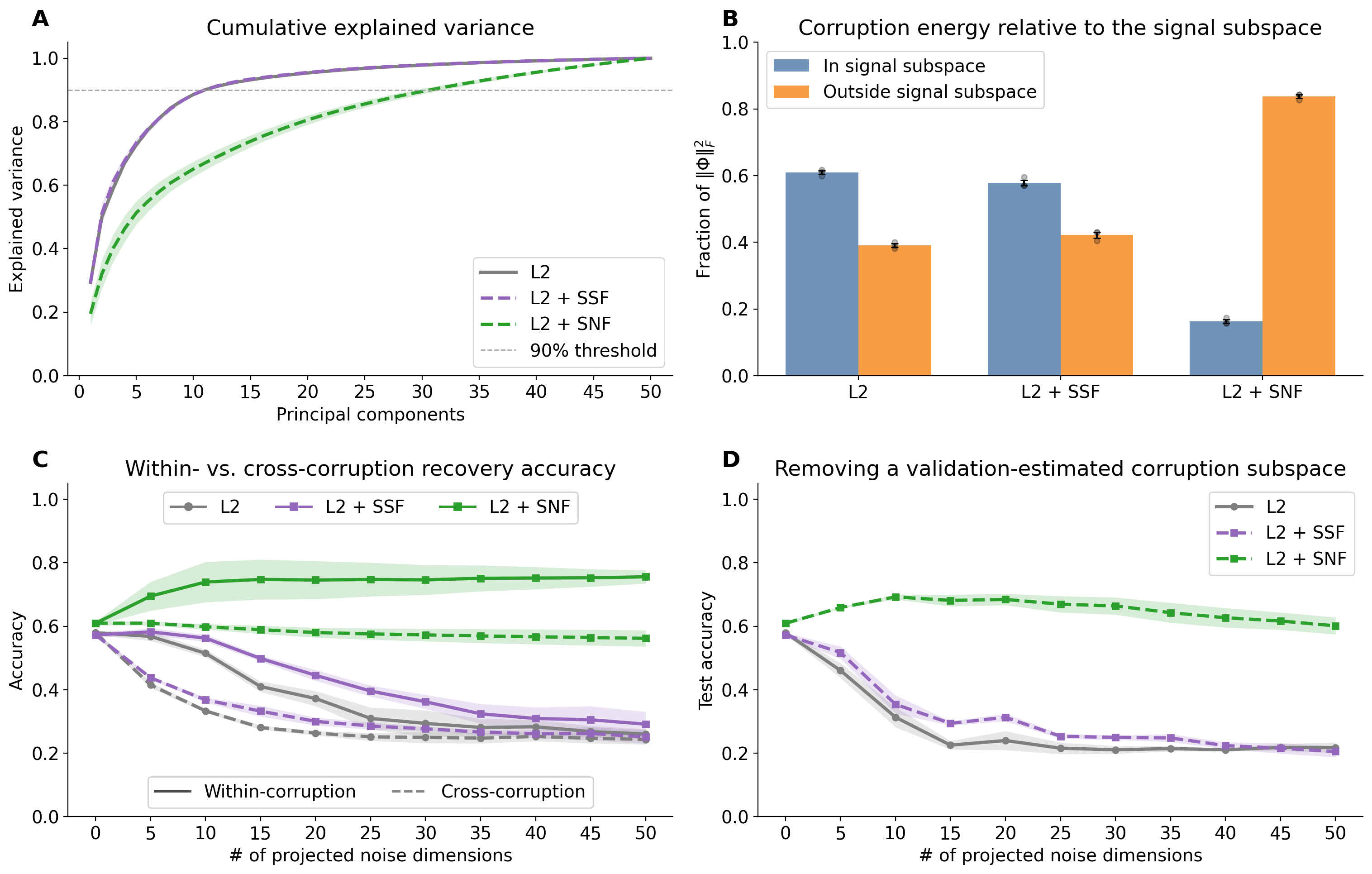}
    \caption{\textbf{SNF-regularized models evict nuisance variation into a separable subspace.} A) Cumulative explained variance of corruption-induced perturbations. B) Fraction of corruption-induced variance falling inside vs.\ outside the signal subspace, by regularizer. C) Within- vs.\ cross-corruption recovery accuracy when the noise subspace is estimated and evaluated on the same test data. D) Test accuracy after projecting out a noise subspace estimated on an independent validation split, demonstrating the subspace generalizes beyond the data used to identify it.}
    \label{fig:nuisance_subspace}
\end{figure}

To determine why SNF regularization produces better performance on corrupted examples, we carried out a mechanistic analysis of each model's representation structure.

Fig.~\ref{fig:nuisance_subspace}a shows the cumulative variance of the perturbation matrix $\Delta$ explained by an increasing number of principal components. The $L_2$ and $L_2$+SSF regularized models are nearly indistinguishable: corruption-induced variance concentrates rapidly, with 11 components sufficient to exceed the 90\% threshold. The $L_2$+SNF regularized model requires substantially more components (33) to reach the same threshold, indicating that SNF regularization causes corruption perturbations to occupy a higher-rank subspace ($r_{\text{eff}} = $11.38$\pm$0.26 versus 11.17$\pm$0.67 and 24.75$\pm$2.48 for $L_2$, $L_2$+SSF, and $L_2$+SNF, respectively). At first, this is counterintuitive, as a regularizer intended to suppress noise appears instead to spread it out.

Surprisingly, this higher rank reflects nuisance variance being separated from the signal subspace, not a failure to suppress noise. The $\mathcal{L}_s$ regularizer (Eq.~\ref{eq:snf}) penalizes noise-subspace variance specifically within the signal subspace $\Phi$, placing no constraint on variance elsewhere. To test whether the elevated rank under SNF regularization reflects this asymmetry, we decomposed corruption-induced variance into components inside and outside the signal subspace (App.~\ref{app:subspace-analysis}). For $L_2$ and $L_2$+SSF regularization, corruption-induced variance is split 61.0\%/39.1\% and 57.9\%/42.2\% between the signal subspace and its complement (Fig.~\ref{fig:nuisance_subspace}b), indicating that a majority of nuisance variation remains entangled with class-discriminative directions under both regimes. Under $L_2$+SNF regularization, this relationship inverts: only 16.3\% of corruption-induced variance falls within the signal subspace, with the remaining 83.7\% displaced into its orthogonal complement. This asymmetry indicates that SNF regularization does not suppress corruption-induced variance globally. It instead moves that variance out of the signal subspace, where it is free to occupy whatever structure is present in the remaining, unconstrained directions of activation space. The elevated rank in Fig.~\ref{fig:nuisance_subspace}a is therefore evidence that successful eviction leaves noise occupying a higher-dimensional region of the orthogonal complement, which carries no constraint on its own rank.

To test whether the nuisance variation subspace could be identified and removed, we projected out the top-$k$ singular vectors of $\Delta$ from each corrupted representation and reclassified the test samples using the frozen linear head, sweeping $k \in \{0, 5, \dots, 50\}$ (Fig.~\ref{fig:nuisance_subspace}c). Estimating and evaluating this subspace on the same corrupted test data yields substantial recovery for $L_2$+SNF regularized model (plateauing at 76.43\% by $k=10$) but not for $L_2$ or $L_2$+SSF regularized ones, whose accuracy instead declines with increasing $k$, consistent with Fig.~\ref{fig:nuisance_subspace}c. This estimate is optimistic, however, since part of the subspace is fit to the same images used to evaluate it. To test whether the identified subspace instead reflects a stable property of the model, we re-estimated the top-$k$ singular vectors of $\Delta$ on the corrupted validation split and applied this fixed subspace to project noise out of test-set representations before classification (Fig.~\ref{fig:nuisance_subspace}d). For $L_2$+SNF regularized model, this validation-estimated projection improves test accuracy from 60.9$\pm$0.6\% at $k=0$ to 69.3$\pm$0.8\% at $k=10$, an improvement of $\sim$8.4\% ($\sim$13.8\% relative), before gradually declining at larger $k$. The smaller performance recovery here, relative to the test-fit estimate above, is expected given the independence of the two splits, and is the more meaningful of the two figures, demonstrating that the noise subspace identified for the $L_2$+SNF regularized model is a property of that transfers across independent samples, rather than an estimation artifact. For $L_2$ and $L_2$+SSF regularized models, the analogous validation-estimated projection does not recover performance and instead degrades it further with increasing $k$, consistent with the entangled geometry indicated by panels b and c.

Together, these results indicate that SNF regularization produces representations in which corruption-induced noise is actively displaced from the signal subspace into a separate, identifiable subspace that is largely corruption-specific (Fig.~\ref{fig:mechanistic_analysis_heatmap}). We also find that projecting out the nuisance variation can reliably improve performance on unseen, corrupted samples.

\section{Conclusion}

In this work, we bridged the regularization-interpretability literature with explicit mechanistic regularizers that directly target geometric properties of learned representations. We showed that $L_2$ regularization operates through multiple distinct mechanisms, a finding reflected in our own experiments and consistent with prior literature on weight decay's diverse effects \citep{neyshabur_search_2014, gunasekar_implicit_2017, arora_implicit_2019, cao_benign_2022, kou_benign_2023}. Explicitly regularizing for signal-signal factorization (SSF) did not yield consistent performance benefits. In contrast, regularizing for signal-noise factorization (SNF) produced meaningful gains, particularly under distribution shift, a benefit that stems from SNF regularization's ability to orthogonalize nuisance variation away from signal-carrying directions, enabling substantial performance recovery on out-of-distribution samples. Together, these results identify a mechanism by which explicit signal-noise factorization can be directly targeted for improving out-of-distribution robustness.

\subsection{Broader impacts}
These findings carry practical relevance beyond benchmark robustness. Machine learning performance in many domains is constrained by nuisance variation: SNF regularization seems to be particularly useful in these situations. In medical imaging, models trained at one hospital often generalize poorly to others, since hospital-specific scanner and staining artifacts become entangled with the diagnostic signal \citep{zech_variable_2018}. A regularizer that explicitly separates the two offers a path toward diagnostic tools that generalize more reliably across clinical sites.  

\subsection{Limitations and future work}
We note several limitations of this work. First, our comparisons are limited to a small set of regularizers as we have not benchmarked against every possible regularization regime. Second, all experiments use a single architecture (ResNet-18), and we cannot say with certainty whether these effects hold for architectures with substantially different inductive biases, such as vision transformers or architectures without residual connections. Third, our evaluation spans only two datasets. We believe that this limitation is at least partially mitigated by CIFAR-100 and BloodMNIST representing quite different domains (natural object recognition versus medical microscopy) such that consistent trends across both offer some initial evidence of generality. Nevertheless, broader validation across additional datasets, modalities, and distribution-shift types is an important avenue for future research.

Future work should explore several directions. First, we tested SSF and SNF regularization independently. Jointly applying both may reveal interactions that neither term captures alone, particularly given the coupling we observed between the two penalties even under independent regularization. Second, our experiments focus on supervised classification. Extending SNF regularization to representation learning settings, such as self-supervised or contrastive pretraining, may be especially promising given that these methods already rely on shaping representation geometry, and explicit signal-noise factorization could offer a principled complement to existing objectives like variance-covariance regularization. Third, the asymmetry we observe where $L_2$ shapes both factorization properties but the reverse does not hold warrants theoretical analysis to characterize precisely which components of $L_2$'s effect are captured by SSF/SNF geometry and which reflect genuinely distinct mechanisms.

\section*{AI use statement}

In this work, we used generative AI tools (Claude, Anthropic) for generating and refining code. We have not used generative AI tools for draft writing but have used them for editing manuscript text and updating/editing figures. Literature search was aided via undermind.ai. Ideation was completed without AI assistance. We have reviewed all AI-assisted work. All AI-generated and AI-refined code was verified by code review before functional validation against expected outputs. We take responsibility for the final content of this work, including text, claims, or artifacts produced with the aid of generative AI.

\section*{Ethics statement}

This work does not involve human subjects, user studies, dataset releases, or applications in sensitive domains. We do not identify any concerns related to bias, fairness, privacy, or security arising from this research. No conflicts of interest or external sponsorship relevant to this work exist. We believe this submission raises no significant ethical concerns beyond those inherent to standard machine learning research.

\section*{Reproducibility Statement}

We provide full mathematical definitions of both regularizers, and their relationship to the theoretical quantities $f$ and $s$ they are designed to target, in Section~\ref{methods:regularizer_def}. Architecture and training details for the CIFAR-100 and BloodMNIST experiments, including hyperparameter search ranges and selected values, software versions, and determinism settings, are provided in App.~\ref{app:ext_methods}. Our approach to statistical reporting given the small number of seeds ($n=3$) is described in Appendix~\ref{app:stats}. Dataset processing steps, including the CIFAR-100-C and MedMNIST-C corruption protocols, are detailed in App.~\ref{app:cifar100c-corr} and App.~\ref{app:medmnistc-corr}. Code to reproduce all experiments and figures is available on github and the link will be added upon acceptance/rejection.

\section*{Author Contributions}

SK: Conceptualization, methodology, investigation, visualization, writing -- original draft. 

JZ: Conceptualization, supervision, writing -- review and editing, funding acquisition.

\section*{Acknowledgments}

This work was supported by Grant $\#$ R01-EY036594 from the National Institutes of Health, an Unrestricted grant from Research to Prevent Blindness to the Jules Stein Eye Institute, Grant $\#$ FA9550-26-1-B113 from the Air Force Office of Scientific Research, a CIFAR Fellow Award for Learning in Machines and Brains, and the Joan and Jerome Snyder Chair in Vision Science at UCLA.

\bibliography{iclr2027_conference}
\bibliographystyle{iclr2027_conference}

\newpage
\appendix
\section{Appendix}

% reset figure counter for appendix
\counterwithin{figure}{section}
\counterwithin{table}{section}
\setcounter{figure}{0}
\setcounter{table}{0}

\subsection{Extended results and figures}

\begin{figure}[htbp]
    \centering
    \includegraphics[width=\linewidth]{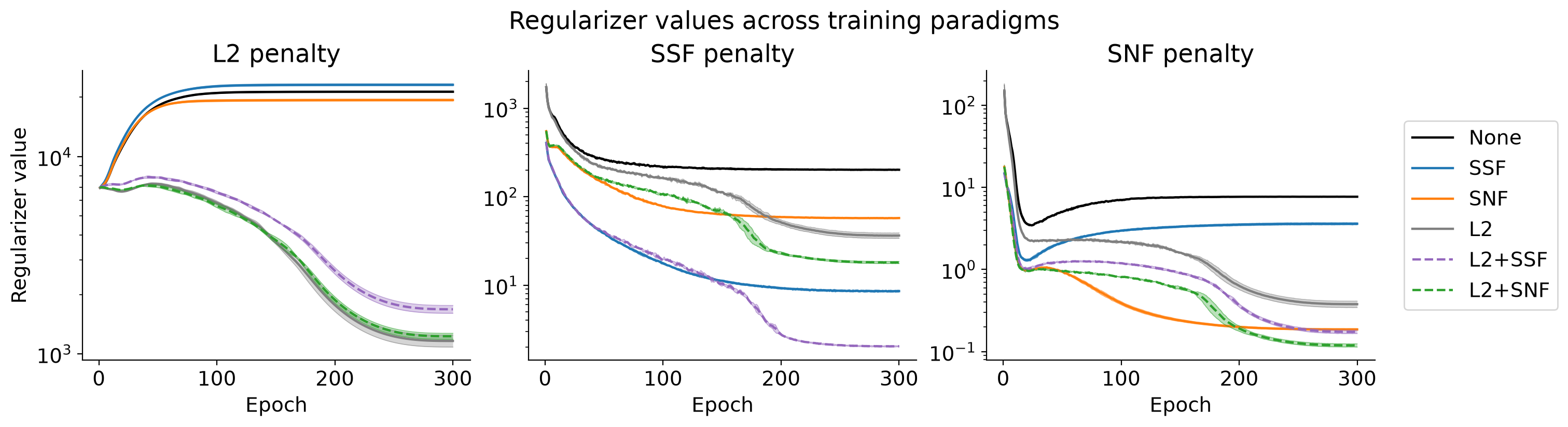}
    \caption{\textbf{Factorization-based regularizer minimization.}
    A) $L_2$ loss term across all training regimes. B) $L_f$ loss term across all training regimes. C) $L_s$ loss term across all training regimes.}
    \label{fig:main_regularizer_minimization}
\end{figure}

\begin{figure}[hbtp]
    \centering
    \includegraphics[width=\linewidth]{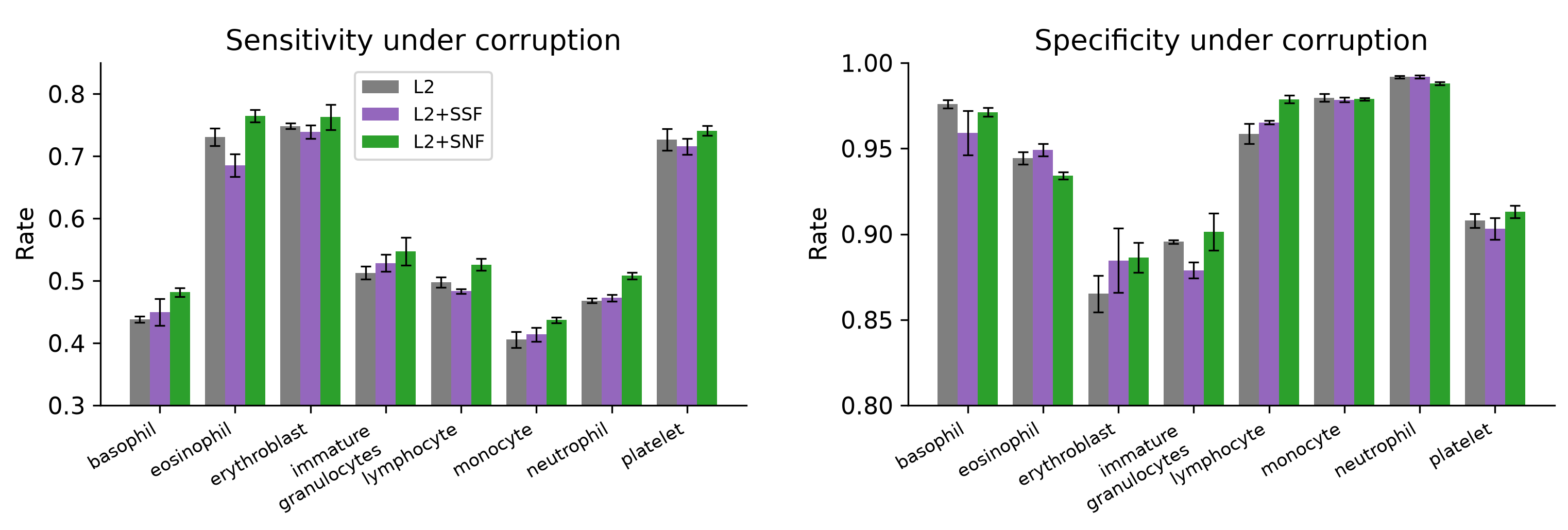}
    \caption{\textbf{Sensitivity and specificity of BloodMNIST dataset by class.}}
    \label{fig:BloodMNIST_sens_spec}
\end{figure}

\begin{figure}[htbp]
    \centering
    \includegraphics[width=\linewidth]{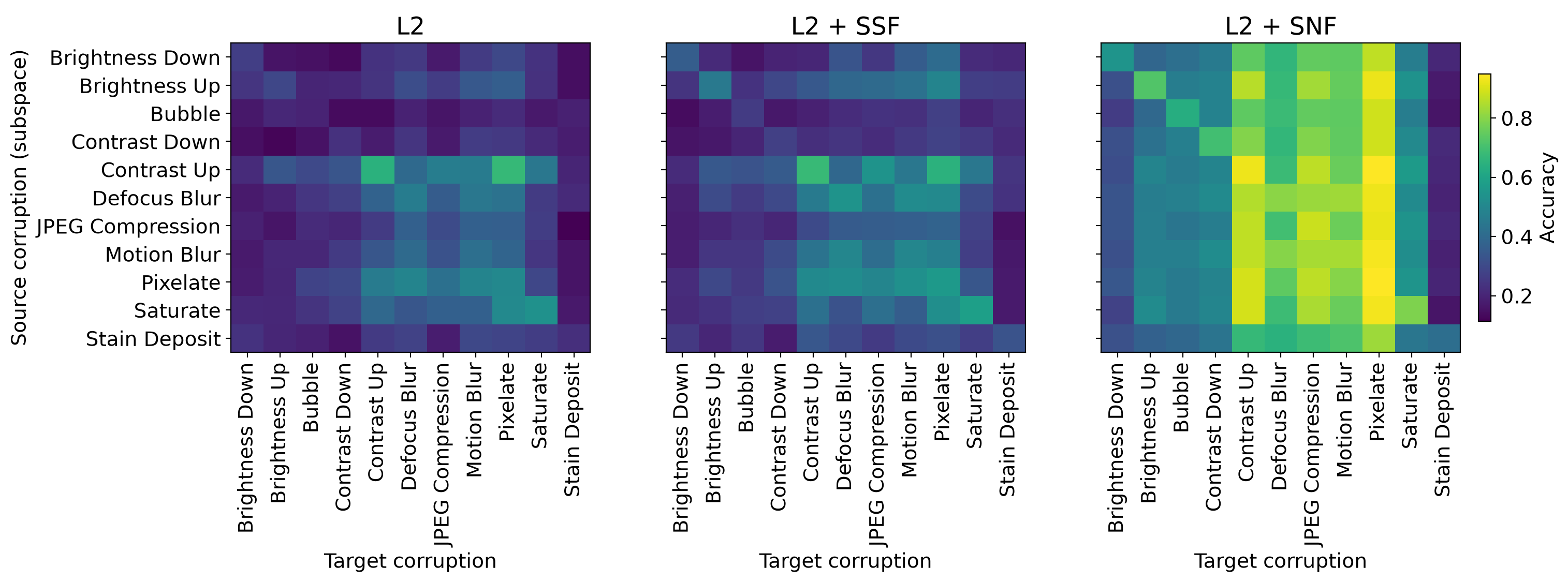}
    \caption{\textbf{Breakdown of performance by corruption type projected out}. For each regularization regime, Y shows the corruption type projected out, X shows the corruption type tested on. Each cell demonstrates the performance on each corruption type after a particular corruption representation is projected out.}
    \label{fig:mechanistic_analysis_heatmap}
\end{figure}

\newpage
\subsection{Extended Methods}
\label{app:ext_methods}

\subsubsection{Architecture details}
For the CIFAR-100 task, we use the standard torchvision ResNet-18 backbone with randomly initialized weights. The final fully connected layer contains 100 output neurons for CIFAR-100's 100 output classes. For the BloodMNIST task, we use 64$\times$64 input images and confirm that the final fully connected layer contains 8 output dimensions for the 8 classes. We train both models from random initialization.

We apply $L_2$ regularization to model weights across all layers of models. The SNF regularizer is applied to the penultimate layer, the 512-dim activation immediately preceding the final fully connected layer. An additional BatchNorm layer (with learnable $\gamma$ and $\beta$) is inserted immediately before the SNF penalty is applied. This layer is present in all conditions, including the unregularized and $L_2$-only baselines, and should be understood as a fixed architectural choice rather than a component of the regularization scheme.

\subsubsection{Statistical reporting}
\label{app:stats}
All models were trained with $n=3$ random seeds (1, 2, 3), with results reported as mean$\pm$standard deviation unless otherwise noted. Given the small number of seeds, we perform no formal statistical tests in this work and avoid inferential language in the text. Instead we rely on per-seed paired differences, which we treat as more informative than a p-value at this sample size.

\subsubsection{CIFAR-100 training details}
\label{app:cifar-training}
Models were trained with stochastic gradient descent (momentum 0.9) and a CosineAnnealing learning rate schedule \citep{loshchilov_sgdr_2017}. Because the SNF regularizer requires inverting the per-batch label covariance $\Omega$, and therefore requires every class to be present in every batch, we use a StratifiedBatchSampler with batch size 1350. Learning rate was scaled to account for the increased batch size, from 0.0579 to 0.579 \citep{goyal_accurate_2018}. Training used an expanded set of augmentations including Pad(4, reflect), RandomHorizontalFlip(p=0.5), RandomRotation(degrees=15), ColorJitter(brightness=0.4, contrast=0.4, saturation=0.4), CIFAR-100-specific normalization, RandomErasing(p=0.5, scale=(0.02, 0.33), ratio=(0.3, 3.3), value=0). Models were trained on the UCLA HPC cluster using 1 NVIDIA RTX A6000 GPU to accommodate large-batch training.

\subsubsection{CIFAR-100 augmentation test}
To test whether our regularizers allow transfer of nuisance variation across augmentation strengths, we train models using two mutually exclusive augmentation strengths (the strong augmentation ranges do not include the weak augmentation ranges) and test the models on augmentations of the opposite strength. To this end, we train the models using the fixed hyperparameters identified in App.~\ref{app:hyperparameters} and use the following augmentations:
\begin{table}[htbp]
    \centering
    \begin{tabular}{lll}
        \hline
        \textbf{Augmentation} & \textbf{Weak range} & \textbf{Strong range} \\
        \hline
        brightness & (0.8, 1.25) & (0.5, 2.0) \\
        contrast & (0.8, 1.25) & (0.5, 2.0) \\
        saturation & (0.8, 1.25) & (0.5, 2.0) \\
        hue & (-0.05, +0.05) & (-0.2, +0.2) \\
        \hline
    \end{tabular}
\end{table} \\
We also use the standard CIFAR-100 augmentations: reflect-pad by 4px, random 32×32 crop, and random horizontal flip.

\subsubsection{BloodMNIST training details}
Model training used the same setup as the original MedMNIST paper: 100 epochs, batch size 112 (original MedMNIST uses 128, we use 112 to satisfy the StratifiedBatchSampling objective), and learning rate $1\times10^{-3}$ with multi-step decay to $1\times10^{-4}$ at epoch 50 and $1\times10^{-5}$ at epoch 75 \citep{yang_medmnist_2023}. We used an $L_2$ weight of $1\times10^{-6}$, the minimum required to reach performance equivalent to the original MedMNIST results. Training included the augmentations used in the MedMNIST baseline, and consisted of RandomHorizontalFlip(p=0.5), RandomVerticalFlip(p=0.5), RandomRotation(degrees=90), RandomResizedCrop(64, scale=(0.8,1.0)). Models were trained on the UCLA HPC cluster using one NVIDIA RTX 2080Ti GPU.

\subsubsection{Hyperparameter search}
\label{app:hyperparameters}
For each dataset (CIFAR-100 and BloodMNIST), hyperparameters were searched individually rather than jointly. We first swept $L_2$ strength alone on the unregularized network and selected the value that maximized validation accuracy. This value was then held fixed while separately sweeping SNF regularization strength (for $L_2$+SNF) and SSF regularization strength (for $L_2$+SSF), so as to isolate the benefit each regularizer confers on top of the selected $L_2$ baseline. On CIFAR-100, we additionally evaluated SNF and SSF regularization in isolation, with $L_2$ strength fixed at 0, to assess their effect independent of weight decay. All selections were made by validation accuracy, with the test set held out until final evaluation. Full search ranges and selected values for both datasets are given in Tables~\ref{tab:hyperparams_cifar} and~\ref{tab:hyperparams_bloodmnist}.

\begin{table}[htbp]
    \centering
    \begin{tabular}{llllll}
        \hline
        \textbf{Configuration} & \textbf{Swept} & \textbf{Held fixed} & \textbf{Search range} & \textbf{Selected} & \textbf{Val Acc} \\
        \hline
        SSF & $\lambda_f$ & $\lambda_{L_2}=0$ & [$10^{-5}$, $10^{-2}$], log & $10^{-4}$ & 0.696$\pm$0.014 \\
        SNF & $\lambda_s$ & $\lambda_{L_2}=0$ & [$10^{-4}$, $10^{-1}$], log & $10^{-2}$ & 0.716$\pm$0.009 \\
        $L_2$ & $\lambda_{L_2}$ & -- & [$10^{-5}$, $10^{-2}$], log & $10^{-4}$ & 0.722$\pm$0.007 \\
        $L_2$+SSF & $\lambda_f$ & $\lambda_{L_2}=10^{-4}$ & [$10^{-5}$, $10^{-2}$], log & $10^{-4}$ & 0.721$\pm$0.005 \\
        $L_2$+SNF & $\lambda_s$ & $\lambda_{L_2}=10^{-4}$ & [$10^{-4}$, $10^{-1}$], log & $10^{-2}$ & 0.730$\pm$0.005 \\
        \hline
    \end{tabular}
    \caption{\textbf{CIFAR-100 hyperparameter search.} Details of ranges searched, values held fixed, and selected hyperparameter values according to validation set performance for the CIFAR-100 task.}
    \label{tab:hyperparams_cifar}
\end{table}

\begin{table}[htbp]
    \centering
    \begin{tabular}{llllll}
        \hline
        \textbf{Configuration} & \textbf{Swept} & \textbf{Held fixed} & \textbf{Search range} & \textbf{Selected} & \textbf{Val Acc} \\
        \hline
        $L_2$ & $\lambda_{L_2}$ & -- & [$10^{-6}$, $10^{-2}$], log & $10^{-6}$ & 0.984$\pm$0.002 \\
        $L_2$+SSF & $\lambda_{f}$ & $\lambda_{L_2}=10^{-6}$ & [$10^{-5}, 10^{-2}$], log & $10^{-3}$ & 0.976$\pm$0.004 \\
        $L_2$+SNF & $\lambda_s$ & $\lambda_{L_2}=10^{-6}$ & [$10^{-4}$, $10^{-1}$], log & $10^{-1}$ & 0.981$\pm$0.001 \\
        \hline
    \end{tabular}
    \caption{\textbf{BloodMNIST hyperparameter search.} Details of ranges searched, values held fixed, and selected hyperparameter values according to validation set performance for the BloodMNIST task.}
    \label{tab:hyperparams_bloodmnist}
\end{table}

\subsubsection{CIFAR-100-C corruptions}
\label{app:cifar100c-corr}
We utilize the pre-prepared CIFAR-100-C dataset, available at https://zenodo.org/records/3555552, described by \citet{hendrycks_benchmarking_2019}. In this work, we use the corrupted dataset to compute model geometry measurements.

\subsubsection{BloodMNIST and MedMNIST-C corruptions}
\label{app:medmnistc-corr}
The BloodMNIST dataset is a sub-dataset of the MedMNIST dataset, a commonly used set of datasets for medical imaging and classification tasks. This dataset contains 8 blood cell types that are pre-cropped and pre-processed. The dataset contains a total of 17,092 images, pre-split into training (11,959), validation (1,712), and testing (3,421) which we kept consistent in this study. The BloodMNIST dataset also contains considerable class imbalance. The details of the dataset breakdown are given in Table~\ref{tab:bloodmnist_data_breakdown}.
\begin{table}[htbp]
\centering
  \begin{tabular}{llll}
      \hline
      \textbf{Class Name} & \textbf{Label} & \textbf{Count (total 17,092)} & \textbf{Percentage} \\
      \hline
      Basophil & 0 & 852 & 7.12\% \\
      Eosinophil & 1 & 2181 & 18.24\% \\
      Erythroblast & 2 & 1085 & 9.07\% \\
      Immature granulocytes & 3 & 2026 & 16.94\% \\
      Lymphocyte & 4 & 849 & 7.10\% \\
      Monocyte & 5 & 993 & 8.30\% \\
      Neutrophil & 6 & 2330 & 19.48\% \\
      Platelet & 7 & 1643 & 13.74\% \\
      \hline
  \end{tabular}
  \caption{\textbf{BloodMNIST dataset breakdown}}
  \label{tab:bloodmnist_data_breakdown}
\end{table}

MedMNIST-C corruptions were applied only to the test set using the `medmnistc' python package's (https://pypi.org/project/medmnistc/) standard implementation of each corruption type across the package's standard 5 severity levels.
\subsubsection{Mechanistic subspace analysis}
\label{app:subspace-analysis}

\paragraph{Representation extraction.} We analyzed the 512-dimensional penultimate-layer activations $Z \in \mathbb{R}^{N \times 512}$, taken from the batch-normalization layer immediately preceding the linear classifier. This layer is the direct target of the SNF penalty during training and is therefore the natural site at which to test its geometric effect. Representations were extracted from the BloodMNIST test set ($N = 3{,}421$ images) under clean conditions and under each of the 55 corruption variants (11 corruption types $\times$ 5 severity levels), using the same ImageNet normalization pipeline applied at training time. All inference was performed with the model in evaluation mode and without gradient computation. Results are reported for seeds 1, 2, and 3 independently and aggregated as mean $\pm$ standard deviation.

\paragraph{Perturbation matrix.} We defined the per-sample perturbation
\begin{equation}
  \Delta_{t,s} = Z_{t,s} - Z_{\text{clean}} \;\in\; \mathbb{R}^{N \times 512}
\end{equation}
where $Z_{\text{clean}} \in \mathbb{R}^{N \times 512}$ denotes clean representations and $Z_{t,s}$ denotes representations under corruption type $t$ at severity $s$. Stacking all $\Delta_{t,s}$ across the 55 variants yields a global perturbation matrix of shape $55N \times 512$.

We additionally decomposed corruption-induced variance into components inside and outside the signal subspace $\Phi$ (Sec.~\ref{sec:prelim}), using the orthogonal projector
\begin{equation}
    P_\Phi = \Phi(\Phi^\top\Phi)^{-1}\Phi^\top,
\end{equation}
\begin{equation}
  \|\Delta_i\|^2 = \underbrace{\|P_\Phi \Delta_i\|^2}_{\text{in-subspace}} +
  \underbrace{\|(I - P_\Phi) \Delta_i\|^2}_{\text{out-of-subspace}},
    \label{eq:corr_decomp}
\end{equation}
where $\Delta_i$ denotes the perturbation of an individual sample $i$.

\paragraph{Noise subspace estimation.} We mean-centered the stacked perturbation matrix and applied a randomized truncated SVD \citep{halko_finding_2010} (top-50 components) to obtain right singular vectors $V \in \mathbb{R}^{512 \times 50}$ and singular values $\sigma_i$. The cumulative variance fraction captured by the top-$k$ directions was computed as
\begin{equation}
  \text{CVF}(k) = \frac{\sum_{i=1}^{k} \sigma_i^2}{\sum_{i=1}^{50} \sigma_i^2}.
\end{equation}
Effective rank was quantified via the exponential of the spectral entropy,
\begin{equation}
  r_{\text{eff}} = \exp\!\left( -\sum_{i} p_i \log p_i \right),
  \qquad p_i = \frac{\sigma_i^2}{\sum_j \sigma_j^2},
\end{equation}
such that a higher $r_{\text{eff}}$ indicates that corruption-induced variance is distributed across more directions in representation space. Note that both quantities are normalized over the retained top-50 components rather than over the full 512-dimensional spectrum, and should therefore be read as describing the distribution of variance within the leading perturbation subspace.

\paragraph{Recovery experiment.} To test whether the noise subspace is separable from the classifier's signal subspace, we orthogonally projected the top-$k$ global singular vectors out of each corrupted representation,
\begin{equation}
  \tilde{Z}_i = Z_i - V_k V_k^\top Z_i,
  \label{eq:proj}
\end{equation}
where $Z_i \in \mathbb{R}^{512}$ denotes a single sample's corrupted representation and $V_k \in \mathbb{R}^{512 \times k}$ contains the first $k$ right singular vectors. The projected representation $\tilde{Z}_i$ was then passed directly to the frozen linear classification head (no retraining of any parameters) and accuracy was evaluated over all $N$ test images. We swept $k \in \{0, 5, 10, 15, 20, 25, 30, 35, 40, 45, 50\}$, where $k = 0$ corresponds to the unmodified corrupted baseline. An increase in accuracy with $k$ indicates that the projected directions carry noise orthogonal to the decision-relevant signal. A decrease indicates entanglement of noise and signal.

\paragraph{Per-corruption transfer experiment.} To assess whether each corruption type occupies a distinct noise subspace, we estimated per-corruption subspaces independently. For source corruption $t$, we stacked $\Delta_{t,s}$ across all five severities to form a matrix of shape $5N \times 512$ and extracted the top-$k$ right singular vectors $V_{t,k}$. We then applied this source subspace to suppress noise in every target corruption $t'$ via the projection in Equation~\ref{eq:proj}, and measured mean accuracy over severities $s \in \{0,\ldots,4\}$. This yields an $11 \times 11$ transfer matrix at each value of $k$, where entry $(t, t')$ is the mean accuracy when the subspace of corruption $t$ is used to denoise corruption $t'$. Diagonal entries (within-corruption) reflect recovery under the matched subspace and off-diagonal entries (cross-corruption) reflect recovery under a mismatched subspace. A large within-cross gap indicates that each corruption type occupies a distinct, non-transferable noise subspace.

\end{document}